\documentclass[11pt]{article}

\usepackage[final]{acl}

\usepackage{times}
\usepackage{latexsym}
\usepackage{array}

\usepackage{float}
\restylefloat{table}

\usepackage[T1]{fontenc}

\usepackage[utf8]{inputenc}

\usepackage{microtype}

\usepackage{inconsolata}

\usepackage{graphicx}

\usepackage[utf8]{inputenc}
\usepackage{arabtex}
\usepackage{utf8}
\setcode{utf8}
\usepackage[utf8]{inputenc}

\title{Multimodal Arabic Captioning with Interpretable Visual Concept Integration}

\author{Passant Elchafei \\
  Ulm University, Germany \\
  \texttt{passant.elchafei@uni-ulm.de} \\\And
  Amany Fashwan \\
  Alexandria University, Egypt\\
  \texttt{amany.fashwan@alexu.edu.eg}\\}

\usepackage{textcomp}      
\usepackage{upquote}       
\usepackage{epstopdf-base} 

\DeclareGraphicsExtensions{.pdf,.png,.jpg,.jpeg}

\begin{document}
\maketitle
\title{Multimodal Arabic Captioning with Interpretable Visual Concept Integration}
\maketitle

\begin{abstract}
We present VLCAP, an Arabic image captioning framework that integrates CLIP-based visual label retrieval with multimodal text generation. Rather than relying solely on end-to-end captioning, VLCAP grounds generation in interpretable Arabic visual concepts extracted with three multilingual encoders, mCLIP, AraCLIP, and Jina V4, each evaluated separately for label retrieval. A hybrid vocabulary is built from training captions and enriched with about 21K general domain labels translated from the Visual Genome dataset, covering objects, attributes, and scenes. The top-$k$ retrieved labels are transformed into fluent Arabic prompts and passed along with the original image to vision–language models. In the second stage, we tested Qwen-VL and Gemini Pro Vision for caption generation, resulting in six encoder–decoder configurations. The results show that mCLIP + Gemini Pro Vision achieved the best BLEU-1 (5.34\%) and cosine similarity (60.01\%), while AraCLIP + Qwen-VL obtained the highest LLM-judge score (36.33\%). This interpretable pipeline enables culturally coherent and contextually accurate Arabic captions.
\end{list}
\end{abstract}

\section{Introduction}
In today's digital age, images are everywhere, shared across social media platforms, embedded in news articles, used in education, e-commerce, and communication. With the rapid growth of visual content, images have become a dominant form of information exchange and expression. This widespread presence highlights the need for intelligent systems that can understand, interpret, and describe visual content effectively \cite{gendy2024advancements}. 

Image captioning is the task of automatically generating syntactically correct and semantically meaningful sentences that describe an image’s content. It plays a vital role in bridging the gap between visual content and natural language. Equipping machines with the ability to interpret and describe visual information offers numerous benefits, including enhanced information retrieval, support for early childhood education, assistance for visually impaired individuals, and applications in social media, among others. While understanding the content of an image may seem effortless, even for children, it remains a significant challenge for computers \cite{eljundi2020resources}. 

Vision-language (VL) models have significantly advanced image understanding and captioning tasks in English and other high-resource languages \cite{zhang2024vision}. However, Arabic image captioning remains underexplored, particularly for culturally rich datasets requiring grounded and interpretable visual understanding. End-to-end generation models often fail to capture the fine-grained semantics, contextual nuances, and socio-cultural cues inherent in Arabic visual scenes.\\
This paper presents our submission to the ImageEval Shared Task, specifically the Image Captioning Models Evaluation Subtask. The objective of this subtask is to develop Arabic image captioning models capable of generating culturally relevant and contextually accurate image descriptions. 
To address this task, we propose VLCAP, a modular Arabic captioning pipeline that integrates visual label reasoning with multimodal generation. Unlike prior work, which mainly adapts English datasets or relies on end-to-end models, our approach explicitly grounds captioning in Arabic visual concepts. In the first stage, we conduct three separate experiments using CLIP-based encoders AraCLIP \cite{al2024araclip}, mCLIP \cite{chen2023mclip}, and Jina V4 \cite{gunther2025jina} to extract the top-$k$ Arabic visual labels for each image. These labels act as interpretable anchors representing “what is seen.”

In the second stage, we use these extracted labels to construct enriched prompts, which are then combined with the original images and fed into two different vision–language models in separate experiments: Qwen‑VL \cite{bai2023qwenvl} and Gemini Pro Vision \cite{google2023gemini}. This setup enables us to systematically evaluate which combination of label extraction model and caption generation model yields the most culturally aligned, semantically accurate, and fluent Arabic captions. By decoupling visual recognition from linguistic description, VLCAP improves both cultural relevance and model transparency.

The rest of this paper is organized as follows: Section \ref{sec:Background} reviews the background and related work on Arabic image captioning and the dataset used in our study, while Section \ref{sec:sysOverview} presents VLCAP, our proposed Arabic vision–language captioning system, and describes the system overview. Section \ref{sec:results} discusses the results of our experiments. Finally, Section \ref{sec:conclusion} concludes the paper.

\begin{figure*}
  \centering
  \includegraphics[width=0.8\textwidth,height=4cm]{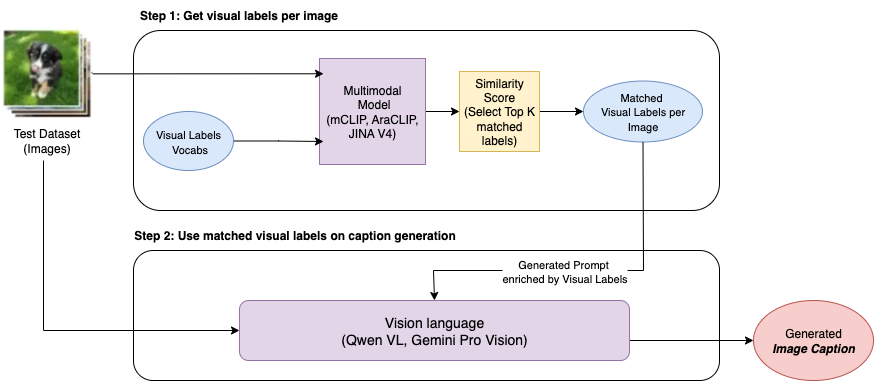}
  \caption{
    \textbf{VLCAP system overview.} \small The framework operates in two stages: (1) Arabic visual labels are retrieved by computing image–text similarity with a multilingual multimodal encoder (mCLIP, AraCLIP, or Jina V4) against a curated label vocabulary; (2) the retrieved labels are inserted into an Arabic prompt, which together with the original image, is passed to a vision–language model (Qwen-VL or Gemini Pro Vision) to generate the final caption.
    }

  \label{fig:framework}
\end{figure*}

\begin{figure*}
  \centering
  \includegraphics[width=0.4\textwidth,height=2.5cm]{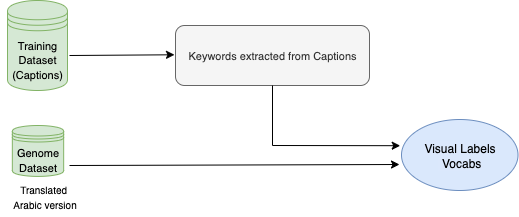}
  \caption{
   \textbf{Visual Labels Vocab Builder}. \small Construction of the Arabic visual label vocabulary: Most frequent content words extracted from the training captions and augmented with general-domain visual concepts translated from the Visual Genome dataset, producing the final vocabulary used for label retrieval. 
  }
  \label{fig:vocab_builder}
\end{figure*}

\begin{table*}[ht]
\centering
\begin{tabular}{|l|c|c|c|}
\hline
\textbf{Model} & \textbf{BLEU-1 Mean} & \textbf{Cosine Similarity Mean} & \textbf{LLM Judge Score} \\
\hline
Base Model \cite{bashiti2025imageeval}   & \textbf{16.98} & 58.46 & 30.82\\ \hline
Gemini+mCLIP & 5.34 & \textbf{60.01}& 33.05\\ \hline
Gemini+AraCLIP & 4.25 & 58.89& \textbf{36.33}\\ \hline 
Gemini+Jina V4 & 4.49 & 57.81 & 34.80\\  \hline
Qwen+mCLIP & 5.20 & 58.39 & 23.49\\ \hline 
Qwen+AraCLIP & 4.57 & 57.19 & 31.40\\ \hline 
Qwen+Jina V4 & 4.17 & 57.03 & 30.35\\
\hline
\end{tabular}
\caption{Performance comparison of our CLIP-augmented captioning system (Gemini and Qwen combined with mCLIP, AraCLIP, and Jina V4) against the Base Model.}
\label{tab:clip_vlm_results}
\end{table*}

\begin{table}[h]
\centering
\begin{tabular}{|p{5.0cm}|p{2.0cm}|}
\hline
\textbf{Participant} & \textbf{Cosine Similarity Mean} \\ \hline
Base Model               & 58.46 \\ \hline
VLCAP (\textbf{Ours})                   &\textbf{ 60.01} \\ \hline
Averroes \cite{averroes2025}                & 58.55 \\ \hline
phantom\_troupe \cite{phantomtroupe2025}          & 57.48 \\ \hline
ImpactAi \cite{ImpactAi2025}                 & 56.22 \\ \hline
Codezone Research Group \cite{asbellobichi2025}  & 38.30 \\ \hline
\end{tabular}
\caption{Cosine Similarity Mean scores for participating teams.}
\label{tab:cosine-similarity}
\end{table}

\begin{table}[h]
\centering
\begin{tabular}{|l|c|}
\hline
\textbf{Participant} & \textbf{LLM Judge Score} \\ \hline
Base Model              & 30.82 \\ \hline
Averroes                     & \textbf{33.97} \\ \hline
VLCAP (\textbf{Ours})    & 33.05 \\ \hline
phantom\_troupe         & 31.43 \\ \hline
ImpactAi                & 26.55 \\ \hline
Codezone Research Group & 15.14 \\ \hline
\end{tabular}
\caption{LLM Judge Score results for participating teams.}
\label{tab:llm-judge-score}
\end{table}

\begin{table*}[h]
\centering
\begin{tabular}{|l|c|c|c|c|}
\hline
\textbf{Participant} & \textbf{Cultural Relevance} & \textbf{Conciseness} & \textbf{Completeness} & \textbf{Accuracy} \\ \hline
VLCAP                   & 2.57          & 3.17          & \textbf{2.67} & \textbf{2.97} \\ \hline
Averroes                     & \textbf{3.63} & \textbf{3.43} & 2.60 & 2.80 \\ \hline
Phantom Troupe          & 3.40          & 3.27          & 2.33 & 2.40 \\ \hline
Codezone Research Group & 1.10          & 2.03          & 1.47 & 2.03 \\ \hline
ImpactAi                & 3.13          & 2.73          & 1.77 & 1.97 \\ \hline
\end{tabular}
\caption{Manual evaluation results based on Cultural Relevance, Conciseness, Completeness, and Accuracy.}
\label{tab:manual-eval}
\end{table*}

\section{Background}\label{sec:Background}
The ImageEval 2025 Shared Task focuses on evaluating image captioning models for the Arabic language, with two main subtasks: (1) Building an open-source dataset of images with culturally appropriate, naturally written Arabic captions, supporting the development of Arabic-native image captioning resources and (2) Automatic generation of Arabic captions for given images. We participated in Subtask 2: Image Captioning Models Evaluation, which requires participants to generate captions for a set of images in Arabic \cite{bashiti2025imageeval}. 

In Subtask 2, the input is a single image, and the output is a short, descriptive caption in \textbf{\textit{Arabic}}. Captions are submitted in CSV format, with each row containing the image\_id and the generated caption. The generated captions are evaluated against a hidden reference set using a combination of automatic metrics ROUGE, BLEU and LLM-as-a-judge to assess semantic similarity and overall quality. The images cover a broad range of everyday scenes, enabling the models to learn both literal and contextually enriched descriptions. The dataset is culturally relevant, reflecting both Modern Standard Arabic and occasional dialectal expressions.

The dataset used in this shared task is entirely in Arabic and consists of high-quality, manually curated captions. It includes a training set of 2,718 images with human-authored captions, a validation set of 76 images with undisclosed gold-standard captions, and a test set of 753 images, also with undisclosed gold-standard captions.

Research on Arabic image captioning has been steadily growing, although it still lags behind progress in English captioning. Early work in the field primarily involved adapting English datasets by translating captions into Arabic or creating Arabic versions of existing datasets like Flickr8k and MS-COCO. \cite{al2018automatic} developed an Arabic image captioning dataset and implemented a model using a convolutional neural network (CNN) for image feature extraction and a recurrent neural network (RNN) with an LSTM decoder for sentence generation. These models laid the foundation for subsequent developments in Arabic captioning. \cite{eljundi2020resources} proposed an end-to-end model that directly transcribes images into Arabic text. They developed an annotated dataset for Arabic image captioning (AIC). They also developed a base model for AIC that relies on text translation from English image captions. The two models are evaluated with the new dataset, and the results show the superiority of their end-to-end model.

More recent studies have focused on building specialized datasets and improving model architectures. \cite{s23083783} built `ArabicFashionData' dataset, which contains labeled images of clothing items. Using this data, researchers developed an attention-based encoder-decoder model that achieved a high BLEU-1 score of 88.52. \cite{za2022bench} highlighted the lack of standardized Arabic benchmarks and proposed unified datasets for evaluating multi-task learning approaches using pre-trained word embeddings, which showed moderate improvements in caption quality.

Transformer-based models have also gained traction. \cite{emami2022arabic} explored Arabic image captioning using deep bidirectional transformers by integrating pre-trained language models into the generation process. \cite{alsayed2023performance} expanded on this by analyzing the impact of text preprocessing tools like CAMeL Tools and various image encoders such as ResNet152. Their experiments demonstrated substantial improvements in BLEU-4 scores up to 148\% and their best-performing model outperformed existing approaches by 379\%.

Vision-language models have introduced further advancements. The VIOLET model \cite{mohamed2023violet} combines a vision encoder with a Gemini text decoder. It leverages an automated method to collect Arabic caption data from English sources, resulting in strong performance, including a CIDEr score of 61.2 on a manually annotated Arabic dataset.

In another recent approach, \cite{elbedwehy2023improved} experimented with combining visual features extracted from powerful image encoders like SWIN, ConvNeXt, and XCIT, alongside Arabic pre-trained language models such as CAMeLBERT and MARBERTv2. This feature fusion strategy significantly improved the fluency and accuracy of the generated captions, outperforming earlier models.

\section{System Overview}\label{sec:sysOverview}
Unlike prior work that mainly adapts English datasets or relies on end-to-end models, our approach explicitly grounds captioning in Arabic visual concepts. We present \textbf{VLCAP}, a modular Arabic image captioning framework that decouples visual label extraction from caption generation to enhance cultural alignment, semantic accuracy, and interpretability. The system operates in two main stages as shown in Figure \ref{fig:framework}, supported by a comprehensive Arabic visual vocabulary constructed through the Visual Label Builder (Figure~\ref{fig:vocab_builder}).

\subsection{Arabic Visual Vocabulary Construction}
As a preliminary step, we built a vocabulary of Arabic visual labels from two sources: \textbf{(1)} most frequent content words extracted from the training captions after removing Arabic stopwords, numbers, and very short tokens, and \textbf{(2)} an augmented set of 21{,}000 high-frequency visual concepts, covering objects, attributes, and scenes, translated from the Visual Genome dataset \cite{krishna2017visual} and adapted to Arabic cultural contexts. This vocabulary serves as the foundation for all subsequent label matching.

\subsection{Visual Label Extraction}
During inference (Figure~\ref{fig:framework}), for each input image, multimodal similarity scores are computed between image embeddings and vocabulary label embeddings using three CLIP-based encoders: AraCLIP, mCLIP, and Jina V4. Cosine similarity ranking is applied to select the top-$k$ matched labels per image, which serve as interpretable visual representations of ``what is seen.'' These matched labels are stored for later use in caption generation. Three separate experiments, one for each CLIP-based encoder, are conducted to determine the most effective label extractor.

\subsection{Prompt-Guided Caption Generation}
The top-ranked labels (typically 25–30) are used to construct an Arabic prompt of the form:
\<باستخدام العناصر التالية> [top-ranked selected labels]  \<محتوى الصورة بدقة.>

This prompt, along with the input image, is fed into a vision–language model. We experiment with two models: Qwen-VL and Gemini Pro Vision. The resulting captions are grounded in both the visual content and the matched labels, ensuring cultural relevance and semantic accuracy.

\subsection{Combination Analysis Evaluation}
By pairing each of the three label extractors with both caption generation models, we evaluate a total of six configurations. The evaluation focuses on identifying the optimal combination for producing culturally aligned, semantically accurate, and fluent Arabic captions.

\section{Results}\label{sec:results}
The official evaluation of shared task submissions employed three complementary methods: Cosine Similarity, which quantifies lexical closeness between generated and reference captions after Arabic-specific normalization and TF–IDF $n$-gram comparison; LLM-as-a-Judge, using OpenAI’s GPT-4o to assess semantic accuracy, relevance, and fluency under reproducible conditions; and Manual Evaluation, where 5\% of the test set was human-rated on cultural relevance, conciseness, completeness, and accuracy.

The results in Table \ref{tab:clip_vlm_results} demonstrate that our system, which integrates CLIP-based visual label detection with Qwen and Gemini, consistently outperforms the base model \cite{bashiti2025imageeval} across all evaluation metrics. While the base model achieves the highest BLEU-1 score due to its direct captioning pipeline, it lags behind in semantic similarity and human-preference evaluations. For the Gemini model, mCLIP yields the strongest BLEU-1 and cosine similarity means, reflecting closer alignment with ground-truth captions and semantic coherence. Notably, AraCLIP, despite lower BLEU-1 and cosine similarity scores, achieves the highest LLM Judge Score, indicating that captions generated with AraCLIP labels are often judged as more contextually relevant or human-preferred. Jina V4 provides balanced performance across all metrics for Gemini. For Qwen, mCLIP again ranks highest in BLEU-1 and cosine similarity, but its LLM Judge Score is relatively low, suggesting that Qwen’s outputs are less favored by human evaluators compared to Gemini. Conversely, AraCLIP and Jina V4 improve Qwen’s LLM Judge Scores, highlighting the role of the CLIP-based label extractor in shaping user-perceived caption quality. 

Overall, these results confirm that our CLIP-augmented system enhances performance beyond the baseline, particularly in semantic similarity and human preference, with the choice of CLIP model exerting a stronger influence on caption quality than the downstream vision–language model itself.

In this shared task, our system achieved the highest performance in the Cosine Similarity metric, ranking first among all participating teams (Table \ref{tab:cosine-similarity}), and secured second place in the LLM-as-a-Judge evaluation (Table \ref{tab:llm-judge-score}), reflecting strong results in both semantic adequacy and fluency. In the manual evaluation Table \ref{tab:manual-eval}, our captions ranked first in Completeness (2.67\%) and Accuracy (2.97\%), while placing second in Cultural Relevance (2.57\%) and Conciseness (3.17\%). These outcomes underscore the system’s ability to generate Arabic captions that are accurate, comprehensive, and semantically faithful, while maintaining competitive performance in cultural appropriateness and conciseness.

\section{Conclusion}\label{sec:conclusion}
In this work, we introduced VLCAP, a modular Arabic image captioning framework that separates visual label extraction from caption generation to enhance cultural alignment, semantic accuracy, and interpretability. Through six experiments combining three CLIP-based encoders with two vision–language models, we found that Gemini Pro Vision + mCLIP delivered the strongest lexical and semantic performance (\textit{BLEU-1:} 5.34, \textit{cosine similarity:} 60.01), whereas Qwen-VL + AraCLIP achieved the highest LLM-based human-alignment score (36.33). These outcomes demonstrate that VLCAP’s decoupled design allows tuning for different evaluation priorities and provides a transferable approach for culturally aware captioning in other low-resource languages.




\bibliography{acl_latex}
\end{document}